\documentclass[runningheads]{llncs}
\usepackage{graphicx}

\usepackage{tikz}
\usepackage{comment}
\usepackage{amsmath,amssymb} 
\usepackage{color}
\usepackage{graphicx}
\usepackage{amsmath}
\usepackage{amssymb}
\usepackage{booktabs}
\usepackage{multirow}
\usepackage{arydshln}
\usepackage{soul}
\usepackage{listings}
\usepackage{caption}
\usepackage{subcaption}

\usepackage[accsupp]{axessibility}  

\usepackage[capitalize]{cleveref}
\crefname{section}{Sec.}{Secs.}
\Crefname{section}{Section}{Sections}
\Crefname{table}{Table}{Tables}
\crefname{table}{Tab.}{Tabs.}
\newcommand\sbullet[0]{\mathbin{\vcenter{\hbox{\scalebox{1.4}{$\bullet$}}}}}
\newcommand\ssquare[0]{\mathbin{\vcenter{\hbox{\scalebox{0.75}{$\blacksquare$}}}}}

\definecolor{ioite}{RGB}{98, 54, 255}
\definecolor{ruby}{RGB}{224, 32, 32}
\definecolor{greycol}{RGB}{230, 230, 230}
\definecolor{ametyst}{RGB}{182, 32, 224}
\definecolor{pink}{RGB}{213, 0, 115}
\definecolor{amber}{RGB}{250, 100, 0}
\definecolor{amazonite}{RGB}{68, 215, 182}
\definecolor{graphite}{RGB}{109, 114,120}
\definecolor{lightgrey}{RGB}{250, 250, 250}

\begin{document}
\pagestyle{headings}
\mainmatter
\def\ECCVSubNumber{5612}  

\title{Continual 3D Convolutional Neural Networks for \\ Real-time Processing of Videos} 

\titlerunning{Continual 3D Convolutional Neural Networks}
%
\author{Lukas Hedegaard
\and
Alexandros Iosifidis
}
\authorrunning{L. Hedegaard and A. Iosifidis}
%
\institute{Department of Electrical and Computer Engineering, Aarhus University, Denmark
\email{\{lhm,ai\}@ece.au.dk}\\
}
\maketitle

\begin{abstract}

We introduce \textit{Continual} 3D Convolutional Neural Networks (\textit{Co}3D CNNs), a new computational formulation of spatio-temporal 3D CNNs, in which videos are processed frame-by-frame rather than by clip.
In online tasks demanding frame-wise predictions, \textit{Co}3D CNNs dispense with the computational redundancies of regular 3D CNNs, namely the repeated convolutions over frames, which appear in overlapping clips.
We show that \textit{Continual} 3D CNNs can reuse preexisting 3D-CNN weights to reduce the per-prediction floating point operations (FLOPs) in proportion to the temporal receptive field while retaining similar memory requirements and accuracy.
bgithuThis is validated with multiple models on \mbox{Kinetics-400} and Charades with remarkable results: \textit{Co}X3D models attain state-of-the-art complexity/accuracy trade-offs on \mbox{Kinetics-400} with 12.1$-$15.3$\times$ reductions of FLOPs and 2.3$-$3.8\% improvements in accuracy compared to regular X3D models while reducing peak memory consumption by up to 48\%. 
Moreover, we investigate the transient response of \textit{Co}3D CNNs at start-up
and perform extensive benchmarks of on-hardware processing characteristics for publicly available 3D CNNs.
%
%


\end{abstract}
\section{Introduction} \label{sec:introduction}


Through the availability of large-scale open-source datasets such as ImageNet~\cite{russakovsky2015imagenet} and Kinetics~\cite{kay2017kinetics}, \cite{carreira2018kinetics},
deep, over-parameterized Convolutional Neural Networks (CNNs) have achieved impressive results in the field of computer vision.
In video recognition specifically, 3D CNNs have lead to multiple breakthroughs in the state-of-the-art~\cite{carreira2017quo}, \cite{tran2018closer}, \cite{feichtenhofer2019slowfast}, \cite{feichtenhofer2020x3d}. 
Despite their success in competitions and benchmarks where only prediction quality is evaluated, computational cost and processing time remains a challenge to the deployment in many real-life use-cases with energy constraints and/or real-time needs.
To combat this general issue, multiple approaches have been explored.
These include computationally efficient architectures for image~\cite{howard2017mobilenet}, \cite{zhang2018shufflenet}, \cite{mingxing2019efficientnet} 
and video recognition~\cite{kopulku2019resource}, \cite{feichtenhofer2020x3d}, \cite{zhu2020faster},
pruning of network weights~\cite{chen2015compressing}, \cite{han2015deep}, \cite{he2017channel},
knowledge distillation~\cite{hinton2015distilling}, \cite{yim2017gift}, \cite{passalis2018learning},
and 
network quantisation~\cite{hubara2016binarized}, \cite{cai2017quantisation}, \cite{floropoulos2019complete}.

The contribution in this paper is complementary to all of the above. It exploits the computational redundancies in the application of regular spatio-temporal 3D CNNs to a continual video stream in a sliding window fashion (\cref{fig:conv-redundancy}).
This redundancy was also explored recently~\cite{kopuklu2020dissected}, \cite{singh2019recurrent} using specialised architectures. However, these are not weight-compatible with regular 3D CNNs.
We present a weight-compatible reformulation of the 3D CNN and its components as a \mbox{\textit{Continual}} 3D Convolutional Neural Network (\textit{Co}3D CNN).
\textit{Co}3D CNNs process input videos frame-by-frame rather than clip-wise and can reuse the weights of regular 3D CNNs, producing identical outputs for networks without temporal zero-padding. 
Contrary to most deep learning papers, the work presented here needed no training; our goal was to validate the efficacy of converting regular 3D CNNs to Continual CNNs directly, and to explore their characteristics in the online recognition domain.
Accordingly, 
we perform conversions from five 3D CNNs, each at different points on the accuracy/speed pareto-frontier, and evaluate their frame-wise performance.
While there is a slight reduction in accuracy after conversion due to zero-padding in the regular 3D CNNs, a simple network modification of extending the temporal receptive field recovers and improves the accuracy significantly \textit{without} any fine-tuning at a negligible increase in computational cost.
Furthermore, we measure the transient network response at start-up, and perform extensive benchmarking on common hardware and embedded devices to gauge the expected inference speeds for real-life scenarios.
Full source code is available at \texttt{\url{https://github.com/lukashedegaard/co3d}}.

\begin{figure}
    \centering
    \includegraphics[width=0.75\linewidth]{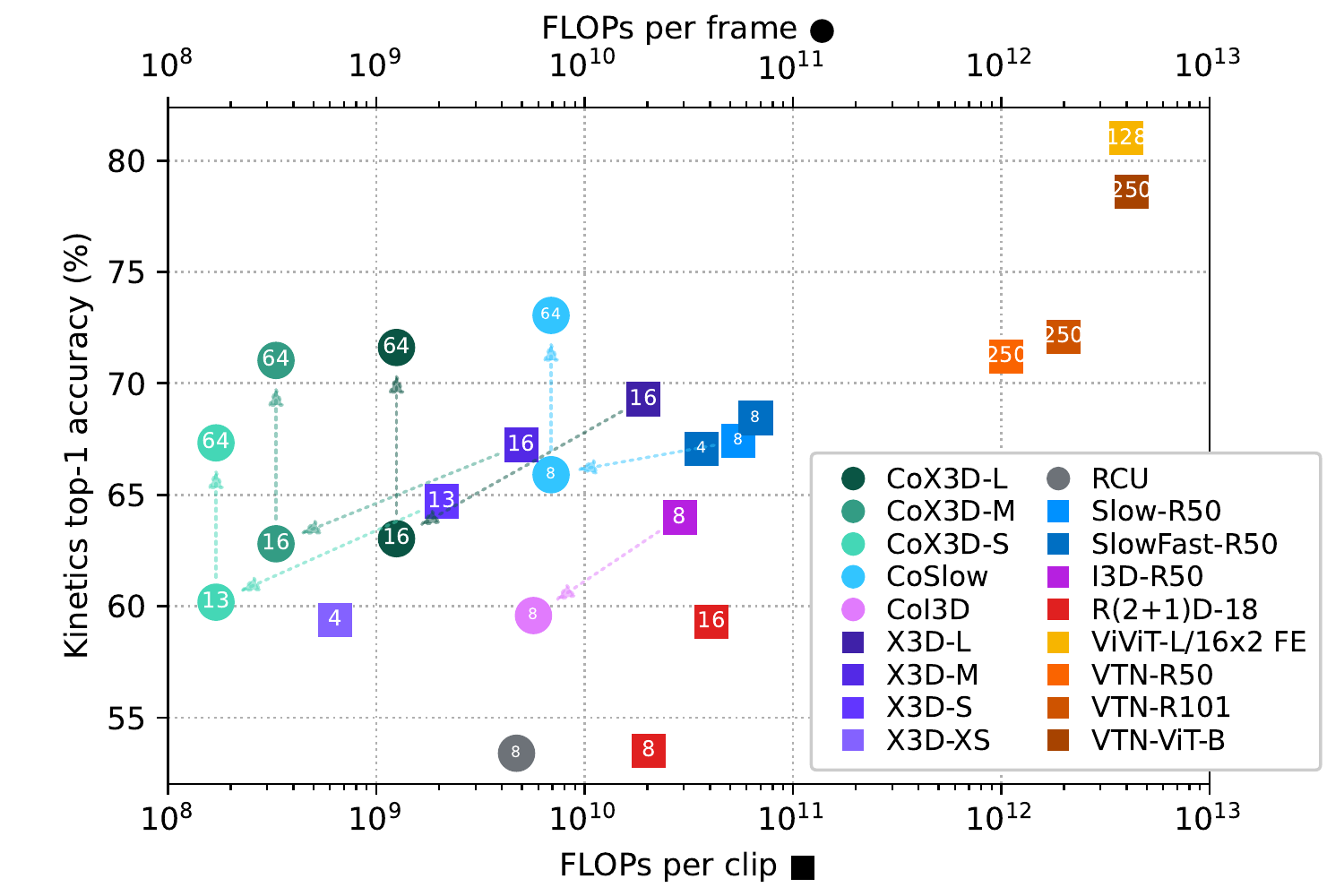}
    \caption{
        \textbf{Accuracy/complexity trade-off} for \textit{Continual} 3D CNNs and recent state-of-the-art methods on Kinetics-400 using 1-clip/frame testing. 
        $\ssquare$~FLOPs per \textit{clip} are noted for regular networks, while $\sbullet$~FLOPs per \textit{frame} are shown for the Continual 3D CNNs. 
        Frames per clip / global average pool size is noted in the representative points.
        Diagonal and vertical arrows indicate a direct weight transfer from regular to Continual 3D CNN and an extension of receptive field.
    }
    \label{fig:test-acc-vs-flops}
\end{figure}
\vspace{-5pt}
\section{Related Works} \label{sec:related-work}
\vspace{-3pt}
\subsection{3D CNNs for video recognition}
\vspace{-3pt}
Convolutional Neural Networks with spatio-temporal 3D kernels may be considered the natural extension of 2D CNNs for image recognition to CNNs for video recognition.
Although they did not surpass their 2D CNN + RNN competitors~\cite{donahue2015longterm}, \cite{yuehei2015beyond} initially~\cite{ji20133dconv}, \cite{karpathy2014largescale}, \cite{tran2015learning}, arguably due to a high parameter count and insufficient dataset size, 3D CNNs have achieved state-of-the-art results on Human Action Recognition tasks~\cite{carreira2017quo}, \cite{tran2018closer}, \cite{feichtenhofer2019slowfast} since the Kinetics dataset~\cite{kay2017kinetics} was introduced.
While recent large-scale Transformer-based methods~\cite{arnab2021vivit}, \cite{neimark2021video} have become leaders in terms of accuracy, 3D CNNs still achieve state-of-the-art accuracy/complexity trade-offs. 
Nevertheless, competitive accuracy comes with high computational cost, which is prohibitive to many real-life use cases.

In image recognition, efficient architectures such as MobileNet~\cite{howard2017mobilenet}, ShuffleNet~\cite{zhang2018shufflenet}, and EfficientNet~\cite{mingxing2019efficientnet} attained improved accuracy-complexity trade-offs.
These architectures were extended to the 3D-convolutional versions {3D-MobileNet} \cite{kopulku2019resource}, {3D-ShuffleNet}~\cite{kopulku2019resource} and X3D~\cite{feichtenhofer2020x3d} ($\approx${3D-EfficientNet}) with similarly improved pareto-frontier in video-recognition tasks. 
While these efficient 3D CNNs work well for offline processing of videos, they are limited in the context of online processing, where we wish to make updates predictions for each frame; real-time processing rates can only be achieved with the smallest models at severely reduced accuracy.
3D CNNs suffer from the restriction that they must process a whole ``clip'' (spatio-temporal volume) at a time. 
When predictions are needed for each frame, this imposes a significant overhead due to repeated computations. 
In our work, we overcome this challenge by introducing an alternative computational scheme for spatio-temporal convolutions, -pooling, and -residuals, which lets us compute 3D CNN outputs frame-wise (continually) and dispose of the redundancies produced by regular 3D CNNs.

\vspace{-3pt}
\subsection{Architectures for online video recognition}
\vspace{-3pt}

A well-explored approach to video-recognition~\cite{donahue2015longterm}, \cite{yuehei2015beyond}, \cite{kalogeiton2017action}, \cite{singh2017online} is to let each frame pass through a 2D CNN trained on ImageNet in one stream alongside a second stream of Optical Flow~\cite{farneback2003twoframe} and integrate these using a recurrent network. 
Such architectures requires no network modification for deployment in online-processing scenarios, lends themselves to caching~\cite{xu2018deepcache}, and are free of the computational redundancies experienced in 3D CNNs.
However, the overhead of Optical Flow and costly feature-extractors pose a substantial disadvantage.

Another approach is to utilise 3D CNNs for feature extraction. 
In \cite{molchanov2016online}, spatio-temporal features from non-overlaping clips are used to train a recurrent network for hand gesture recognition.
In \cite{kopuklu2019yowo}, a 3D CNN processes a sliding window of the input to perform spatio-temporal action detection.
These 3D CNN-based methods have the disadvantage of either not producing predictions for each input frame~\cite{molchanov2016online} or suffering from redundant computations from overlapping clips~\cite{kopuklu2019yowo}.

Massively Parallel Video Networks~\cite{carreira2018massively} split a DNN into depth-parallel sub-networks across multiple computational devices to improve online multi-device parallel processing performance. While their approach treats networks layers as atomic operations and doesn't tackle the fundamental redundancy of temporal convolutions, \textit{Continual} 3D CNNs reformulate the network layers, remove redundancy, and accelerate inference on single devices as well.

Exploring modifications of the spatio-temporal 3D convolution operating frame by frame, the Recurrent Convolutional Unit (RCU)~\cite{singh2019recurrent} replaces the 3D convolution by aggregating a spatial 2D convolution over the current input with a 1D convolution over the prior output. 
Dissected 3D CNNs~\cite{kopuklu2020dissected} (D3D) cache the $1 \times n_H \times n_W$ frame-level features in network residual connections and aggregate them with the current frame features via $2 \times 3 \times 3$ convolutions.
Like the our proposed \textit{Continual} 3D CNNs, both RCU and D3D are causal and operate frame-by-frame.
However, they are speciality architectures, which are incompatible with pre-trained 3D CNNs, and must be trained from scratch.
We reformulate spatio-temporal convolutions in a one-to-one compatible manner, allowing us to reuse existing model weights. 


\vspace{-2mm}
\section{Continual Convolutional Neural Networks}\label{sec:continual}
\vspace{-2mm}
\subsection{Regular 3D-convolutions lead to redundancy}\label{sec:regular-cnn-redundancy}
\begin{figure}
    \centering
    \includegraphics[width=0.8\linewidth]{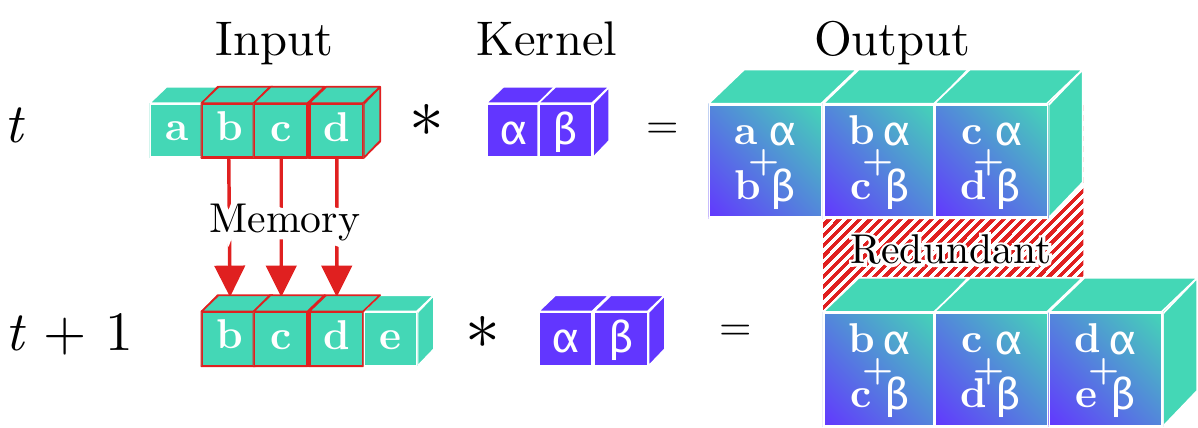}
	\caption{
	\textbf{Redundant computations} for a temporal convolution during online processing, as illustrated by the repeated convolution of inputs (green $\mathbf{b,c,d}$) with a kernel (blue $\alpha, \beta$) in the temporal dimension.
	Moreover, prior inputs ($\mathbf{b,c,d}$) must be stored between time-steps for online processing tasks. }
    \label{fig:conv-redundancy}
    \vspace{-5mm}
\end{figure}

\vspace{-1mm}
Currently, the best performing architectures (e.g., X3D~\cite{feichtenhofer2020x3d} and SlowFast~\cite{feichtenhofer2019slowfast}) employ variations on 3D convolutions as their main building block and perform predictions for a spatio-temporal input volume (video-clip). 
These architectures achieve high accuracy with reasonable computational cost for predictions on clips in the offline setting.
They are, however, ill-suited for online video classification, 
where the input is a continual stream of video frames and a class prediction is needed for each frame. 
For regular 3D CNNs processing clips of $m_T$ frames to be used in this context, prior $m_T-1$ input frames need to be stored between temporal time-steps and assembled to form a new video-clip when the next frame is sampled.
This is illustrated in \cref{fig:conv-redundancy}.

Recall the computational complexity for a 3D convolution:
\begin{equation}
    \Theta( [k_H \cdot k_W \cdot k_T + b] \cdot c_I \cdot c_O \cdot n_H \cdot n_W \cdot n_T ),
\end{equation}
where $k$ denotes the kernel size, $T$, $H$, and $W$ are time, height, and width dimension subscripts, $b \in \{0,1\}$ indicates whether bias is used, and $c_I$ and $c_O$ are the number of input and output channels. 
The size of the output feature map is $n = (m + 2p - d \cdot (k-1) - 1)/s + 1$ for an input of size $m$ and a convolution with padding $p$, dilation $d$, and stride $s$.
%
During online processing, every frame in the continual video-stream will be processed $n_T$ times (once for each position in the clip), leading to a redundancy proportional with $n_T-1$.
%
Moreover, the memory-overhead of storing prior input frames is 
\begin{equation}
    \Theta(c_I \cdot m_H \cdot m_W \cdot [m_T - 1])),
\label{eq:conv-mem-store-frames}
\end{equation}
and during inference the network has to transiently store feature-maps of size 
\begin{equation}
    \Theta(c_O \cdot n_H \cdot n_W \cdot n_T).
    \label{eq:reg-conv-transient}
\end{equation}

\vspace{-3mm}
\subsection{Continual Convolutions}
\begin{figure}
    \centering
    \includegraphics[width=0.8\linewidth]{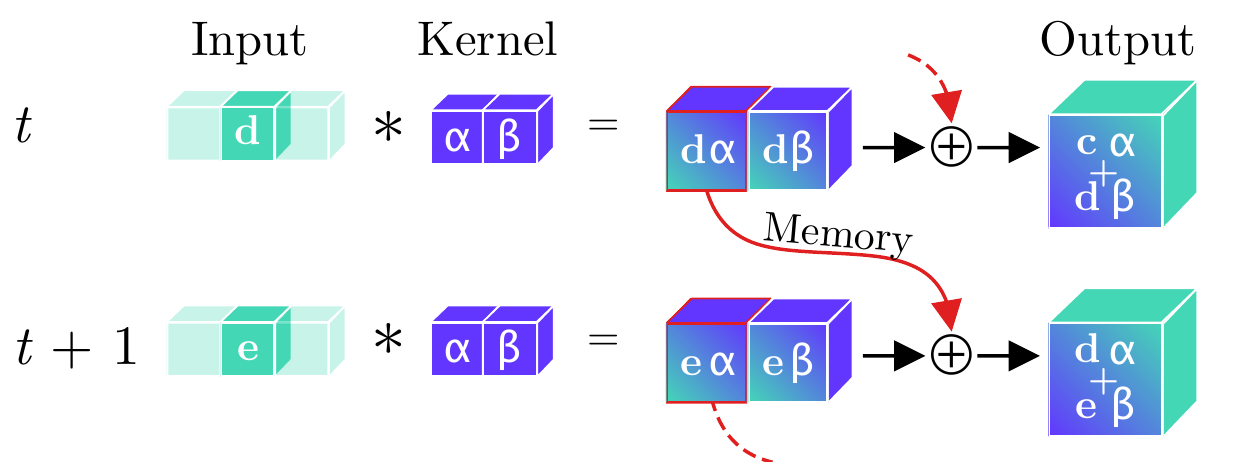}
	\caption{
	\textbf{Continual Convolution}. 
	An input (green $\mathbf{d}$ or $\mathbf{e}$) is convolved with a kernel (blue $\alpha, \beta$). The intermediary feature-maps corresponding to all but the last temporal position are stored, while the last feature map and prior memory are summed to produce the resulting output. For a continual stream of inputs, Continual Convolutions produce identical outputs to regular convolutions.
	}
    \label{fig:conv-fixed}
    \vspace{-5mm}
\end{figure}

\vspace{-1mm}
We can remedy the issue described in \cref{sec:regular-cnn-redundancy} by employing an alternative sequence of computational steps. 
In essence, we reformulate the repeated convolution of a (3D) kernel with a (3D) input-clip that continually shifts along the temporal dimension as a \textit{Continual} Convolution (\textit{Co}Conv), where all convolution computations (bar the final sum) for the (3D) kernel with each (2D) input-frame are performed in one time-step. 
Intermediary results are stored as states to be used in subsequent steps, while previous and current results are summed up to produce the output. 
The process for a 1D input and kernel, which corresponds to the regular convolution in \cref{fig:conv-redundancy}, is illustrated in \cref{fig:conv-fixed}. 
%
In general, this scheme can be applied for online-processing of any $N$D input, where one dimension is a temporal continual stream.
Continual Convolutions are causal~\cite{oord2016wavenet} with no information leaking from future to past 
and can be efficiently implemented by zero-padding the input frame along the temporal dimension with $p = \text{floor}(k / 2)$. 
Python-style pseudo-code of the implementation is shown in \cref{code:coconv3d}.
\begin{center}
\noindent\begin{minipage}{0.8\columnwidth}
\begin{lstlisting}[
    label={code:coconv3d},
    language=python,
    caption={
    \textbf{Pseudo-code} for Continual Convolution.
    Ready-to-use modules are available in the Continual Inference library~\cite{hedegaard2022colib}.
    %and OpenDR~\cite{opendr2022} toolkits.
    }
]
def coconv3d(frame, prev_state = (mem, i)):
    frame = spatial_padding(frame)
    frame = temporal_padding(frame)
    feat = conv3d(frame, weights) 
    output, rest_feat = feat[0], feat[1:]
    mem, i = prev_state or init_state(output)
    M = len(mem)
    for m in range(M):
        output += mem[(i + m) % M, M - m - 1]
    output += bias
    mem[i] = rest_feat
    i = (i + 1) % M
    return output, (mem, i)
\end{lstlisting}
\end{minipage}
\end{center}

In terms of computational cost, we can now perform frame-by-frame computations much more efficiently than a regular 3D convolution. The complexity of processing a frame becomes:
\begin{equation}
    \Theta( [k_H \cdot k_W \cdot k_T + b] \cdot c_I \cdot c_O \cdot n_H \cdot n_W).
\end{equation}
This reduction in computational complexity comes at the cost of a memory-overhead in each layer due to the state that is kept between time-steps. 
The overhead of storing the partially computed feature-maps for a frame is:
\begin{equation}
    \Theta( d_T \cdot [k_T -1] \cdot c_O \cdot n_H \cdot n_W).
\end{equation}
However, in the context of inference in a deep neural network, the transient memory usage within each time-step is reduced by a factor of $n_T$ to
\begin{equation}
    \Theta(c_O \cdot n_H \cdot n_W).
\end{equation}

The benefits of Continual Convolutions thus include the independence of clip length on the computational complexity, state overhead, and transient memory consumption.
The change from (non-causal) regular convolutions to (causal) Continual Convolutions has the side-effect of introducing a delay to the output. 
This is because some intermediary results of convolving a frame with the kernel are only added up at a later point in time (see \cref{fig:conv-fixed}). 
The delay for a continual convolution is
\begin{equation}
    \Theta(d_T \cdot [k_T - p_T - 1]).
\label{eq:coconv-delay}
\end{equation}

\subsection{Continual Residuals} \label{sec:residual}
The delay from Continual Convolutions has an adverse side-effect on residual connections. 
Despite their simplicity in regular CNNs, we cannot simply add the input to a Continual Convolution with its output because the \textit{Co}Conv may delay the output.
Residual connections to a \textit{Co}Conv must therefore be delayed by an equivalent amount (see \cref{eq:coconv-delay}).
This produces a memory overhead of 
\begin{equation}
    \Theta( d_T \cdot [k_T -1] \cdot c_O \cdot m_H \cdot m_W).
\label{eq:residual-mem}
\end{equation}

\vspace{-3mm}
\subsection{Continual Pooling}  \label{sec:pooling}
\vspace{-1mm}
The associative property of pooling operations allows for pooling to be decomposed across dimensions, i.e. $\textrm{pool}_{T, H, W}(\mathbf{X}) = \textrm{pool}_{T}(\textrm{pool}_{H, W}\left(\mathbf{X})\right)$.
For continual spatio-temporal pooling, the pooling over spatial dimensions is equivalent to a regular pooling, while the intermediary pooling results must be stored for prior temporal frames. For a pooling operation with temporal kernel size $k_T$ and spatial output size $n_H \cdot n_W $, the memory consumption is
\begin{equation}
    \Theta([k_T - 1] \cdot n_H \cdot n_W),
\label{eq:pooling-mem}
\end{equation}
and the delay is
\begin{equation}
    \Theta(k_T - p_T - 1).
\label{eq:pooling-delay}
\end{equation}
Both memory consumption and delay scale linearly with the temporal kernel size.
Fortunately, the memory consumed by temporal pooling layers is relatively modest for most CNN architectures ($1.5 \%$ for \textit{Co}X3D-M, see Appendix A). Hence, the delay rather than memory consumption may be of primary concern for real-life applications.
For some network modules it may even make sense to skip the pooling in the conversion to a Continual CNN.
One such example is the 3D Squeeze-and-Excitation (SE) block~\cite{hu2018squeeze} in X3D, where global spatio-temporal average-pooling is used in the computation of channel-wise self-attention. 
Discarding the temporal pooling component (making it a 2D SE block) shifts the attention slightly (assuming the frame contents change slowly relative to the sampling rate) but avoids a considerable temporal delay.


\subsection{The issue with temporal padding}
\begin{figure}[t]
     \centering
     \begin{subfigure}[b]{0.48\linewidth}
         \centering
         \includegraphics[width=0.7\textwidth]{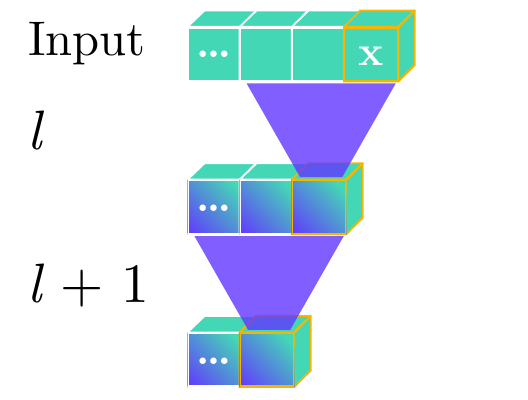}
         \caption{No padding}
         \label{fig:no_pad}
     \end{subfigure}
     \hfill
     \begin{subfigure}[b]{0.48\linewidth}
         \centering
         \includegraphics[width=0.7\textwidth]{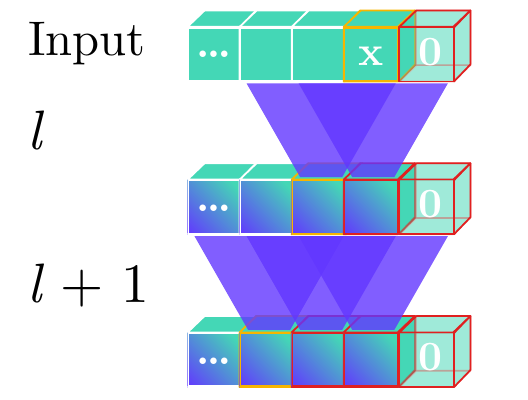}
         \caption{Zero padding}
         \label{fig:padding_issue}
     \end{subfigure}
        \vspace{-5pt}
        \caption{\textbf{Issue with temporal padding}: The latest frame $\mathbf{x}$ is propagated through a CNN with (purple) temporal kernels of size 3 (a) without or (b) with zero padding. Highlighted cubes can be produced only in the latest frame, with yellow boarder indicating independence of padded zero and red boarders dependence. In the zero-padded case (b), the number of frame features dependent on $\mathbf{x}$ following a layer $l$ increases with the number of padded zeros.}
        \label{fig:padding}
        \vspace{-5mm}
\end{figure}

Zero-padding of convolutional layers is a popular strategy for retaining the spatio-temporal dimension of feature-maps in consecutive CNN layers.
For Continual CNNs, however, temporal zero-padding poses a problem, as illustrated in \cref{fig:padding}. 
Consider a 2-layer 1D CNN where each layer has a kernel size of 3 and zero padding of 1.
For each new frame in a continual stream of inputs, the first layer $l$ should produce two output feature-maps: One by the convolution of the two prior frames and the new frame, and another by convolving with one prior frame, the new frame, and a zero-pad. 
The next layer $l+1$ thus receives two inputs and produces three outputs which are dependent on the new input frame of the first layer (one for each input and another from zero-padding).
In effect, each zero padding in a convolution forces the next layer to retrospectively update its output for a previous time-step in a non-causal manner.
Thus, there is a considerable downside to the use of padding.
This questions the necessity of zero padding along the temporal dimension.
In regular CNNs, zero padding has two benefits:
It helps to avoid spatio-temporal shrinkage of feature-maps when propagated through a deep CNN,
and it prevents information at the boarders from ``washing away''~\cite{karpathy2020cs231n}.
The use of zero-padding, however, has the downside that it alters the input-distribution along the boarders significantly~\cite{liu2018partialpadding}, \cite{nguyen2019distribution}. 
For input data which is a continual stream of frames, a shrinkage of the feature-size in the temporal dimension is not a concern, and an input frame (which may be considered a border frame in a regular 3D CNN) has no risk of ``washing away" because it is a middle frame in subsequent time steps.
Temporal padding is thus omitted in Continual CNNs.
As can be seen in the experimental evaluations presented in the following, this constitutes a ``model shift'' in the conversion from regular to Continual 3D CNN if the former was trained with temporal padding.

\vspace{-10pt}

\subsection{Initialisation}\label{sec:init}
Before a Continual CNN reaches a steady state of operation, it must have processed $r_T - p_T - 1$ frames where $r_T$ and $p_T$ are the aggregated temporal receptive field and padding of the network.
For example, Continual X3D-${\{\text{S, M, L}\}}$ models have receptive fields of size $\{69, 72, 130\}$, aggregated padding $\{28, 28, 57\}$, and hence need to process $\{40, 43, 72\}$ frames prior to normal operation.
The initial response depends on how internal state variables are initialised. In \cref{sec:exp-transient}, we explore this further with two initialisation variants: 
1)~Initialisation with \textit{zeros} and 
2)~by repeating a ~\textit{replicate} of the features corresponding to the first input-frame. 
The latter corresponds to operating in a steady state for a ``boring video''~\cite{carreira2017quo} which has one frame repeated in the entire clip.

\subsection{Design considerations} \label{sec:design-considerations}
\vspace{-1mm}
Memory consumption is highly dependent on the clip size employed by the respective models. 
Disregarding the storage requirement of the model weights (which is identical between for regular and continual 3D CNNs), X3D-M has a worst-case total memory-consumption of $7{,}074{,}816$ floats when prior frames and the transient feature-maps are taken into account.
Its continual counterpart, \textit{Co}X3D-M, has a worst case memory only $5{,}072{,}688$ floats. 
How can this be? 
Since Continual 3D CNNs do not store prior input frames and has smaller transient feature maps, the memory savings outweigh the cost of caching features in each continual layer.
Had the clip size been four instead of sixteen, X3D-M$_4$ would have had a worst-case memory consumption of $1{,}655{,}808$ floats and \textit{Co}X3D-M$_4$ of $5{,}067{,}504$ floats, and for clip size of 64, X3D-M$_{64}$ consumes $28{,}449{,}792$ floats and \textit{Co}X3D-M$_{64}$ uses $5{,}093{,}424$ floats.
The memory consumption of regular 3D CNNs is this thus highly dependent on the clip size, while \textit{Co}3D CNNs are not.
%
Continual CNNs utilise longer effective clip sizes much more efficiently than regular CNNs in online processing scenarios. 
In networks intended for embedded systems or online processing, we may thus increase the clip size to achieve higher accuracy with minimal penalty in computational complexity and worst-case memory consumption.

Another design-consideration, which has a considerable influence on memory consumption is the temporal kernel size and dilation of \textit{Co}Conv layers.
Fortunately, the trend to employ small kernel sizes leaves the memory consumption reasonable for current state-of-the-art 3D CNNs~\cite{carreira2017quo}, \cite{tran2018closer}, \cite{feichtenhofer2019slowfast}, \cite{feichtenhofer2020x3d}. A larger temporal kernel size would not only affect the memory growth through the \textit{Co}Conv filter, but also for co-occuring residual connections, since these  consume a significant fraction of the total state-memory for real-life networks; in a Continual X3D-M model (\textit{Co}X3D-M) the memory of residual constitutes $20.5 \%$ of the total model state memory (see Appendix A).

\vspace{-2mm}
\subsection{Training}
\textit{Co}3D CNNs are trained with back-propagation like other neural networks. However, special care must be taken in the estimation of data statistics in normalisation layers: 1) Momentum should be adjusted to $\text{mom}_\text{step} = 2 / (1 + \text{timesteps}\cdot(2 / \text{mom}_\text{clip} - 1))$ to match the exponential moving average dynamics of clip-based training, where T is the clip size; 2) statistics should not be tracked for the transient response.
Alternatively, they can be trained offline in their ``unrolled'' regular 3D-CNN form with no temporal padding. 
This is similar to utilising pre-trained weights from a regular 3D CNN, as we do in our experiments.

\vspace{-2mm}
\section{Experiments} 
\vspace{-2mm}
The experiments in this section aim to show the characteristics and advantages of Continual 3D CNNs as compared with regular 3D CNNs.
One of the main benefits of \textit{Co}3D CNNs is their ability to reuse the network weights of regular 3D CNNs. 
As such, all \textit{Co}3D CNNs in these experiments use publicly available pre-trained network weights of regular 3D CNNs~\cite{feichtenhofer2019slowfast}, \cite{feichtenhofer2020x3d}, \cite{fan2021pytorchvideo} without further fine-tuning.
Data pre-processing follows the respective procedures associated with the originating weights unless stated otherwise.
%
The section is laid out as follows: 
First, we showcase the network performance following weight transfer from regular to Continual 3D on multiple datasets for Human Activity Recognition. 
This is followed by a study on the transient response of \textit{Co}3D CNNs at startup.
Subsequently, we show how the computational advantages of \textit{Co}3D CNNs can be exploited to improve accuracy by extending the temporal receptive field.
Finally, we perform an extensive on-hardware benchmark of prior methods and Continual 3D CNNs, measuring the 1-clip/frame accuracy of publicly available models, as well as their inference throughput on various computational devices.




\begin{table}[t]
\begin{center}
\resizebox{\textwidth}{!}{
\begin{tabular}{llrrrrrrrr}
    \toprule
    &\multirow{2}{*}{\textbf{Model}} 
    &\textbf{Acc.}
    &\textbf{Par.}
    &\textbf{Mem.}  
    &\textbf{FLOPs}
    &\multicolumn{4}{c}{\textbf{Throughput} (preds/s)}
        \\ \cline{7-10}
        && (\%) & (M) & (MB) & (G) & CPU & TX2  & Xavier  &2080Ti
    \\
    \midrule
    \parbox[t]{1mm}{\multirow{9}{*}{\rotatebox[origin=c]{90}{Clip}}}                  
    & I3D-R50                           & 63.98     & 28.04    & 191.59    & 28.61      & 0.93   & 2.54     & 9.20      & 77.15  \\
    & R(2+1)D-18$_8$                    & 53.52     & 31.51    & 168.87    & 20.35      & 1.75   & 3.19     & 6.82      & 130.88 \\
    & R(2+1)D-18$_{16}$                 & 59.29     & 31.51    & 215.44    & 40.71      & 0.83   & 1.82     & 3.77      & 75.81  \\
    & Slow-8×8-R50                      & 67.42     & 32.45    & 266.04    & 54.87      & 0.38   & 1.34     & 4.31      & 61.92  \\
    & SlowFast-8×8-R50                  & 68.45     & 66.25    & 344.84    & 66.25      & 0.34   & 0.87     & 2.72      & 30.72  \\ 
    & SlowFast-4×16-R50                 & 67.06     & 34.48    & 260.51    & 36.46      & 0.55   & 1.33     & 3.43      & 41.28  \\ 
    & X3D-L                             & 69.29     & 6.15     & 240.66    & 19.17      & 0.25   & 0.19     & 4.78      & 36.37  \\
    & X3D-M                             & 67.24     & 3.79     & 126.29    & 4.97       & 0.83   & 1.47     & 17.47     & 116.07 \\
    & X3D-S                             & 64.71     & 3.79     & 61.29     & 2.06       & 2.23   & 2.68     & 42.02     & 276.45 \\
    & X3D-XS                            & 59.37     & 3.79     & 28.79     & 0.64       & 8.26   & 8.20     & 135.39    & 819.87 \\
    \midrule
    \parbox[t]{1mm}{\multirow{11}{*}{\rotatebox[origin=c]{90}{Frame}}} & RCU$_8$~\cite{singh2019recurrent}$^\dagger$     & 53.40     & 12.80     & - & 4.71 & - & - & - & - \\
    & \textit{Co}I3D$_{8}$              & 59.58     & 28.04    & 235.87    & 5.68       & 3.00   & 2.41     & 14.88     & 125.59 \\
    & \textit{Co}I3D$_{64}$             & 56.86     & 28.04    & 236.08    & 5.68       & 3.15   & 2.41     & 14.89     & 126.32 \\
    & \textit{Co}Slow$_{8}$             & 65.90     & 32.45    & 175.98    & 6.90       & 2.80   & 1.60     & 6.18      & 113.77 \\
    & \textbf{\textit{Co}Slow$_{64}$}    & \textbf{73.05}  & \textbf{32.45}  & \textbf{176.41}  & \textbf{6.90}   & \textbf{2.92}   & \textbf{1.60}  & \textbf{6.19} & \textbf{102.00}  \\
    & \textit{Co}X3D-$\text{L}_{16}$    & 63.03     & 6.15     & 184.29    & 1.25       & 2.30   & 0.99     & 25.17     & 206.65 \\
    & \textbf{\textit{Co}X3D-$\text{L}_{64}$}    & \textbf{71.61}     & \textbf{6.15}    & \textbf{184.37}     & \textbf{1.25}      &  \textbf{2.30}   & \textbf{0.99}       & \textbf{27.56}       & \textbf{217.53}  \\
    & \textit{Co}X3D-$\text{M}_{16}$    & 62.80     & 3.79    & 68.88     & 0.33      & 7.57   & 7.26       & 88.79      & 844.73  \\
    & \textbf{\textit{Co}X3D-$\text{M}_{64}$}    & \textbf{71.03}     & \textbf{3.79}    & \textbf{68.96}    & \textbf{0.33}      & \textbf{7.51}   & \textbf{7.04}       & \textbf{86.42}      & \textbf{796.32}  \\
    & \textit{Co}X3D-$\text{S}_{13}$    & 60.18     & 3.79    & 41.91     & 0.17      & 13.16  & 11.06      & 219.64      & 939.72 \\
    & \textbf{\textit{Co}X3D-$\text{S}_{64}$}    & \textbf{67.33}     & \textbf{3.79}    & \textbf{41.99}     & \textbf{0.17}      & \textbf{13.19}  & \textbf{11.13}     & \textbf{213.65}      & \textbf{942.97} \\
    \bottomrule
\end{tabular}
}
\end{center}
\caption{
    \textbf{Kinetics-400 benchmark}. The noted accuracy is the single clip or frame top-1 score using RGB as the only input-modality. 
    The performance was evaluated using publicly available pre-trained models without any further fine-tuning. 
    For speed comparison, predictions per second denote frames per second for the \textit{Co}X3D models and clips per second for the remaining models. Throughput results are the mean of 100 measurements. 
    Pareto-optimal models are marked with bold.
    Mem. is the maximum allocated memory during inference noted in megabytes.
    $^\dagger$Approximate FLOPs derived from paper (see \cref{apx:rcu}).
}
\label{tab:benchmark-kinetics400}
\vspace{-9mm}
\end{table}

\vspace{-2mm}
\subsection{Transfer from regular to Continual CNNs} \label{exp:transfer-reg-to-co}
\vspace{-1mm}
To gauge direct transferability of 3D CNN weights, we implement continual versions of various 3D CNNs and initialise them with their publicly available weights for Kinetics-400~\cite{kay2017kinetics} and Charades~\cite{sigurdsson2016charades}. 
While it is common to use an ensemble prediction from multiple clips to boost video-level accuracy on these benchmarks, we abstain from this, as it doesn't apply to online-scenarios. Instead, we report the single-clip/frame model performance. 

\paragraph{Kinetics-400.}
We evaluate the X3D network variants XS, S, M, and L on the test set using one temporally centred clip from each video. 
The XS network is omitted in the transfer to \textit{Co}X3D, given that it is architecturally equivalent to S, but with fewer frames per clip. 
In evaluation on Kinetics-400, we faced the challenge that videos were limited to 10 seconds. 
Due to the longer transient response of Continual CNNs (see \cref{sec:exp-transient}) and low frame-rate used for training X3D models ($5.0, 6.0, 6.0$ FPS for S, M, and L), the video-length was insufficient to reach steady-state for some models.
As a practical measure to evaluate near steady-state, we repeated the last video-frame for a padded video length of $\approx80\%$ of the network receptive field as a heuristic choice.
The Continual CNNs were thus tested on the last frame of the padded video and initialised with the prior frames. 
The results of the X3D transfer are shown in \cref{tab:benchmark-kinetics400} and \cref{fig:test-acc-vs-flops}.



For all networks, the transfer from regular to Continual 3D CNN results in significant computational savings. 
For the S, M, and L networks the reduction in FLOPs is 12.1$\times$, 15.1$\times$, and 15.3$\times$ respectively.
The savings do not quite reach the clip sizes since the final pooling and prediction layers are active for each frame.
As a side-effect of the transfer from zero-padded regular CNN to Continual CNN without zero-padding, we see a notable reduction in accuracy. 
This is easily improved by using an extended pooling size for the network (discussed in \cref{sec:design-considerations} and in \cref{sec:exp-extended-receptive-fields}). 
Using a global average pooling with temporal kernel size 64, we improve the accuracy of X3D by 2.6\%, 3.8\%, and 2.3\% in the Continual S, M, and L network variants.
As noted, Kinetics dataset did not have sufficient frames to fill the temporal receptive field of all models in these tests.
We explore this further in Sections \ref{sec:exp-transient} and \ref{sec:exp-extended-receptive-fields}.

\vspace{-3mm}
\paragraph{Charades.}
To showcase the generality of the approach, we repeat the above described procedure with another 3D CNN, the \textit{Co}Slow network~\cite{feichtenhofer2019slowfast}. 
We report the video-level mean average precision (mAP) of the validation split alongside the FLOPs per prediction in \cref{tab:charades}. Note the accuracy discrepancy between 30 view (10 temporal positions with 3 spatial positions each) and 1 view (spatially and temporally centred) evaluation. As observed on Kinetics, the \textit{Co}Slow network reduces the FLOPs per prediction proportionally with the original clip size (8 frames), and can recover accuracy by extending the global average pool size.


\begin{table}[t]
	\begin{center}
	\begin{tabular}{llccc}
		\toprule
        & \textbf{Model}                & \textbf{FLOPs (G) × views}        & \textbf{mAP (\%)}  \\
		\midrule
        \parbox[t]{1mm}{\multirow{3}{*}{\rotatebox[origin=c]{90}{Clip}}}
		& Slow-8×8~\cite{feichtenhofer2019slowfast}  
		                                & $54.9 \times 30$          & 39.0      \\
        & Slow-8×8~\cite{feichtenhofer2019slowfast}$^\dagger$     
                                        & $54.9 \times 1$           & 21.4      \\
        & Slow-8×8 (ours)    & $54.9 \times 1$           & 24.1      \\

		\midrule
		\parbox[t]{1mm}{\multirow{2}{*}{\rotatebox[origin=c]{90}{Fr.}}}
        & \textit{Co}Slow$_8$             & $6.9 \times 1$          & 21.5      \\
        & \textit{Co}Slow$_{64}$          & $6.9 \times 1$          & 25.2      \\
        
		\bottomrule
	\end{tabular}
	\end{center}
	\caption{
	    \textbf{Charades benchmark.} Noted are the FLOPs $\times$ views and video-level mean average precision (mAP) on the validation set using pre-trained model weights. 
        $^\dagger$Results achieved using the publicly available SlowFast code~\cite{feichtenhofer2019slowfast}.
	}
	\label{tab:charades}
	\vspace{-10mm}
\end{table}

\renewcommand*{\thefootnote}{\arabic{footnote}}

\vspace{-3mm}
\subsection{Ablation Experiments}
\vspace{-1mm}
As described in \cref{sec:init}, Continual CNNs exhibit a transient response during their up-start.
In order to gauge this response, we perform ablations on the Kinetics-400 validation set, this time sampled at 15 FPS to have a sufficient number of frames available.
This corresponds to a data domain shift~\cite{wang2018deep} relative to the pre-trained weights, where time advances slower.

\vspace{-3mm}
\subsubsection{Transient response of Continual CNNs.} \label{sec:exp-transient}
\begin{figure}[b!]
    \centering
    \begin{subfigure}[b]{0.7\linewidth}
        \centering
        \includegraphics[width=1.0\linewidth]{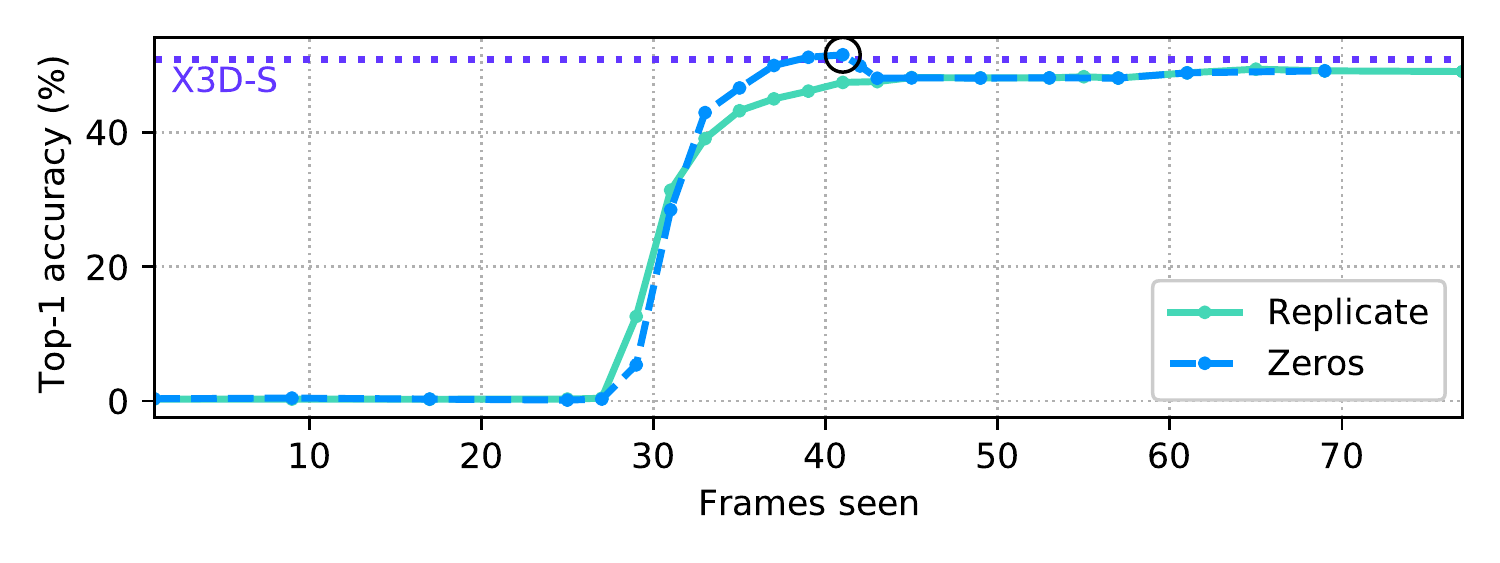}
        \\[-2ex]
        \caption{\textit{Co}X3D-S}
        \label{fig:transient-s}
    \end{subfigure}
    \begin{subfigure}[b]{0.7\linewidth}
        \centering
        \includegraphics[width=1.0\linewidth]{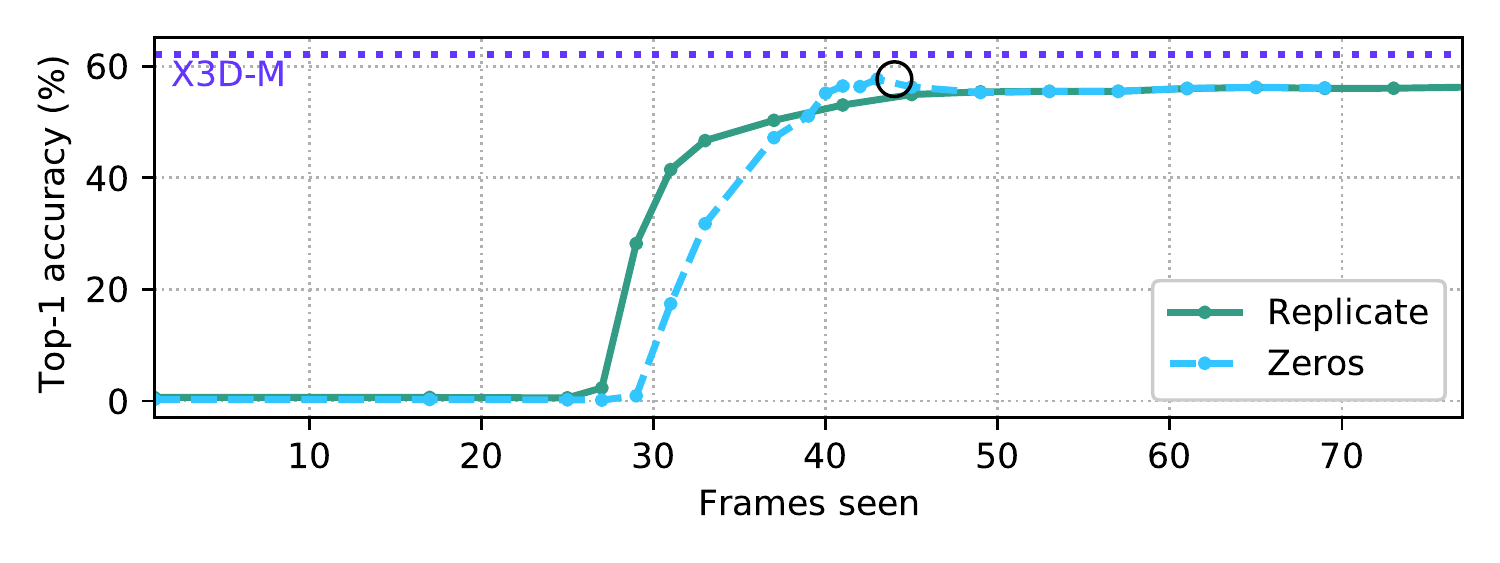}
        \\[-2ex]
        \caption{\textit{Co}X3D-M}
        \label{fig:transient-m}
    \end{subfigure}
    \\[-1ex]
    \setlength{\belowcaptionskip}{-5pt}
	\caption{
	\textbf{Transient response} for Continual X3D-\{S,M\} on the Kinetics-400 val at 15 FPS.
	Dotted horizontal lines denote X3D validation accuracy for 1-clip predictions. 
	Black circles highlight the theoretically required initialisation frames.
	}
    \label{fig:transient_response_results}
    \vspace{-2mm}
\end{figure}

Our expected upper bound is given by the baseline X3D network 1-clip accuracy at 15 FPS.
The transient response is measured by varying the number of prior frames used for initialisation before evaluating a frame using the \textit{Co}X3D model.
Note that temporal center-crops of size $T_{\text{init}} + 1$, where $T_{\text{init}}$ is the number of \text{initialisation frames}, are used in each evaluation to ensure that the frames seen by the network come from the centre. 
This precaution counters a data-bias, we noticed in Kinetics-400, namely that the start and end of a video are less informative and contribute to worse predictions than the central part. 
We found results to vary up to 8\% for a X3D-S network evaluated at different video positions.
The experiment is repeated for two initialisation schemes, ``zeros'' (used in other experiments) and ``replicate'', and two model sizes, S and M.
The transient responses are shown in \cref{fig:transient_response_results}.

For all responses, the first $\approx$25 frames produce near-random predictions, before rapidly increasing at 25$-$30 frames until a steady-state is reached at $49.2\%$ and $56.2\%$ accuracy for S and M.
Relative to the regular X3D, this constitutes a steady-state error of $-1.7\%$ and $-5.8\%$.
Comparing initialisation schemes, we see that the ``replicate'' scheme results in a slightly earlier rise. 
The rise sets in later for the ``zeros'' scheme, but exhibits a sharper slope, topping with peaks of $51.6\%$ and $57.6\%$ at 41 and 44 frames seen as discussed in \cref{sec:init}.
This makes sense considering that the original network weights were trained with this exact amount of zero-padding. Adding more frames effectively replaces the padded zeros and causes a slight drop of accuracy in the steady state, where the accuracy settles at the same values as for the ``replication'' scheme. 

\vspace{-3mm}
\subsubsection{Extended receptive field.} \label{sec:exp-extended-receptive-fields}
Continual CNNs experience a negligible increase in computational cost when larger temporal receptive field are used (see \cref{sec:design-considerations}). 
For \textit{Co}X3D networks, this extension can be trivially implemented by increasing the temporal kernel size of the last pooling layer.
In this set of experiments, we extend \textit{Co}X3D-$\{\text{S,M,L}\}$ to have temporal pooling sizes 32, 64, and 96, and evaluate them on the Kinetics-400 validation set sampled at 15 FPS.
The Continual CNNs are evaluated at frames corresponding to the steady state.

\begin{table}[b!]
	\begin{center}
	
	\begin{tabular}{lllcrr}
		\toprule
		\textbf{Model}  & \textbf{Size} & \textbf{Pool} & \textbf{Acc.} & \textbf{FLOPs (K)} & \textbf{Rec. Field} \\
		\midrule
		\multirow{3}{*}{X3D} 
		                & S             & 13            & 51.0          & 2,061,366 
		                                                                                    & 13                 \\
		                & M             & 16            & 62.1          & 4,970,008 
		                                                                                    & 16                 \\
		                & L             & 16            & 64.1          & 19,166,052 
		                                                                                    & 16                 \\
		\midrule
		    &\multirow{5}{*}{S}         & 13            & 49.2          & 166,565 
		                                                                                    & 69                 \\
		                &               & 16            & 50.1          & 166,567 
		                                                                                    & 72                 \\
		                &               & 32            & 54.7          & 166,574 
		                                                                                    & 88                 \\
		                &               & 64            & 59.8          & 166,587 
		                                                                                    & 120                \\
		                &               & 96            &\textit{61.8}  & 166,601 
		                                                                                    & 152                \\
		                \cline{2-6}
 		\multirow{4}{*}{\textit{Co}X3D}
		    &\multirow{4}{*}{M}         & 16            & 56.3          & 325,456 
		                                                                                    & 72                 \\
		                &               & 32            & 60.7          & 325,463 
		                                                                                    & 88                 \\
		                &               & 64            & 64.9          & 325,477 
		                                                                                    & 120                \\
		                &               & 96            &\textit{67.3 } & 325,491 
		                                                                                    & 152                \\
		                \cline{2-6}
		    &\multirow{4}{*}{L}         & 16            & 53.0          & 1,245,549 
		                                                                                    & 130                \\
		                &               & 32            & 58.5          & 1,245,556 
		                                                                                    & 146                \\
		                &               & 64            &\textit{64.3}  & 1,245,570 
		                                                                                    & 178                \\
		                &               & 96            &\textit{66.3}  & 1,245,584 
		                                                                                    & 210                \\

		\bottomrule
	\end{tabular}
	\end{center}
	\caption{\textbf{Effect of extending pool size}. 
	Note that the model weights were trained at different sampling rates than evaluated at (15 FPS), resulting in a lower top-1 val. accuracy.
	\textit{Italic numbers} denote measurement taken within the transient response due to a lack of frames in the video-clip.
	}
	\label{tab:extended_window_size}
\end{table}

\cref{tab:extended_window_size} shows the measured accuracy and floating point operations per frame (\textit{Co}X3D) / clip (X3D) as well as the pool size for the penultimate network layer (global average pooling) and the total receptive field of the network in the temporal dimension. 
As found in \cref{exp:transfer-reg-to-co}, each transfer results in significant computational savings alongside a drop in accuracy.
Extending the kernel size of the global average pooling layer increases the accuracy of the Continual CNNs by 11.0$-$13.3$\%$ for 96 frames relative the original 13$-$16 frames, surpassing that of the regular CNNs.
Lying at 0.017$-$0.009$\%$, the corresponding computational increases can be considered negligible.

\subsection{Inference benchmarks} \label{sec:inference-benchmarks}
Despite their high status in activity recognition leader-boards~\cite{paperswithcode2021kinetics400},
it is unclear how recent 3D CNNs methods perform in the online setting, where speed and accuracy constitute a necessary trade-off.
To the best of our knowledge, there has not yet been a systematic evaluation of throughput for these video-recognition models on real-life hardware.
In this set of experiments, we benchmark the FLOPs, parameter count, maximum allocated memory and 1-clip/frame accuracy of I3D~\cite{carreira2017quo}, R(2+1)D~\cite{tran2018closer}, SlowFast\cite{sovrasov2020ptflops}, X3D~\cite{feichtenhofer2020x3d}, \textit{Co}I3D, \textit{Co}Slow, and \textit{Co}X3D.
To gauge achievable throughputs at different computational budgets, networks were tested on four hardware platforms as described in Appendix B.

As seen in the benchmark results found in \cref{tab:benchmark-kinetics400},
the limitation to one clip markedly lowers accuracy compared with the multi-clip evaluation published in the respective works~\cite{carreira2017quo}, \cite{tran2018closer}, \cite{feichtenhofer2019slowfast}, \cite{feichtenhofer2020x3d}.
Nontheless, the Continual models with extended receptive fields attain the best accuracy/speed trade-off by a large margin. 
For example, \textit{Co}X3D-$\text{L}_{64}$ on the Nvidia Jetson Xavier achieves an accuracy of 71.3\% at 27.6 predictions per second compared to 67.2\% accuracy at 17.5 predictions per second for X3D-M while reducing maximum allocated memory by 48\%! 
Confirming the observation in \cite{ma2018shufflenetv2}, we find that the relation between model FLOPs and throughput varies between models, with better ratios attained for simpler models (e.g., I3D) than for complicated ones (e.g., X3D). 
This relates to different memory access needs and their cost. Tailor-made hardware could plausibly reduce these differences.
Supplementary visualisation of the results in \cref{tab:benchmark-kinetics400} are found in Appendix C.
\section{Conclusion} \label{sec:conclusion}
We have introduced Continual 3D Convolutional Neural Networks (\textit{Co}3D CNNs), a new computational model for spatio-temporal 3D CNNs, which performs computations frame-wise rather than clip-wise while being weight-compatible with regular 3D CNNs. 
In doing so, we are able dispose of the computational redundancies faced by 3D CNNs in continual online processing, giving up to a 15.1$\times$ reduction of floating point operations,
a 9.2$\times$ real-life inference speed-up on CPU, 48\% peak memory reduction,
and an accuracy improvement of $5.6\%$ on Kinetics-400 through an extension in the global average pooling kernel size.

While this constitutes a substantial leap in the processing efficiency of energy-constrained and real-time video recognition systems, there are still unanswered questions pertaining to the dynamics of \textit{Co}3D CNNs.
Specifically,
the impact of extended receptive fields on the networks ability to change predictions in response to changing contents in the input video is untested.
We leave these as important directions for future work.



\section*{Acknowledgement}
This work has received funding from the European Union’s Horizon 2020 research and innovation programme under grant agreement No 871449 (OpenDR). 

\clearpage
%
%
\bibliographystyle{splncs04}
\bibliography{references}

\clearpage
\appendix
\renewcommand{\thesection}{\Alph{section}}

\DeclareRobustCommand{\highlight}[1]{{\sethlcolor{greycol}\hl{#1}}}
\newcommand{\mathcolorbox}[2]{\colorbox{#1}{$\displaystyle #2$}}


\section*{Appendix}

\section{Worst-case memory for \textit{Co}X3D-M}
In this section, we provide a detailed overview of the memory consumption incurred by the internal state in a Continual X3D-M (\textit{Co}X3D-M) model.
For Continual 3D CNNs, there is no need to store input frames between time steps, though this is the case for regular 3D CNNs applied in an online processing scenario. 
Intermediary computations from prior frames are kept in the continual layers as state if a layer has a temporal receptive field larger than 1. 
A continual $k_T \times k_H \times k_W = 1 \times 3 \times 3$ convolution is equivalent to a regular convolution, while a $3 \times 1 \times 1$ is not. 
The same principle holds for pooling layers. 
As a design decision, the temporal component of the average pooling of Squeeze-and-Excitation (SE) blocks is discarded. Hence, SE blocks do not incur a memory overhead or delay. 
Keeping the temporal pooling of the SE block would have increased memory consumption by a modest $85.050$ ($+1.4\%$).
We can compute the total state overhead using \cref{eq:conv-mem-store-frames}, \cref{eq:residual-mem}, and \cref{eq:pooling-mem} by adding up the state size of each applicable layer shown in \cref{tab:x3d-mem}.
An overview of the resulting computations can be found in \cref{tab:cox3dm-mem-calc}. 
The total memory overhead for the network state is $4{,}771{,}632$ floating point operations.
In addition to the state memory, the worst-case transient memory must be taken into account. 
The largest intermediary feature-map is produced after the first convolution in conv$_1$ and has a size of $24 \times 112 \times 112 = 301{,}056$ floats.
The total worst-case memory consumption for \textit{Co}X3D-M (excluding models weights) is thus \textbf{5,072,688} floats.

If we were to reduce the model clip size from 16 to 4, this would result in a memory reduction of $5{,}184$ floats (only $\text{pool}_5$ is affected) for a total worst-case memory of $5{,}067{,}504$ floats ($-0.1\%$). 
Increasing the clip size to 64 would yield an increased state memory of $20{,}736$ floats giving a total worst-case memory of $5{,}093{,}424$ floats ($+0.4\%$).

\begin{table}
\begin{center}
\begin{tabular}{llrr}
    \toprule
    \textbf{Stage}  & \textbf{Layer}        &                                                               & \textbf{Mem.} (floats) \\ 
    \midrule
    conv$_1$        & conv$_{\text{T}}$     & $(5-1) \times 24 \times 112 \times 112 =$                     & $1{,}204{,}224$ \\
    \midrule
    res$_2$         & residual$_1$          & $(3-1-1) \times 24 \times 112 \times 112 =$                     & $301{,}056$ \\
                    & residual$_{2-3}$      & $\left[(3-1-1) \times 24 \times 56 \times 56\right] \times 2 =$ & $150{,}528$ \\
                    & conv$_{1-3}$          & $\left[(3-1-1) \times 54 \times 56 \times 56\right] \times 3 =$ & $508{,}032$ \\
    \midrule
    res$_3$         & residual$_1$          & $(3-1-1) \times 24 \times 56 \times 56 =$                       & $75{,}264$ \\
                    & residual$_{2-5}$      & $\left[(3-1-1) \times 48 \times 28 \times 28\right] \times 4 =$ & $150{,}528$ \\
                    & conv$_{1-5}$          & $\left[(3-1) \times 108 \times 28 \times 28\right] \times 5 =$  & $846{,}720$ \\
    \midrule
    res$_4$         & residual$_1$          & $(3-1-1) \times 48 \times 28 \times 28 =$                        & $37{,}632$ \\
                    & residual$_{2-11}$     & $\left[(3-1-1) \times 96 \times 14 \times 14\right] \times 10 =$ & $188{,}160$ \\
                    & conv$_{1-11}$         & $\left[(3-1) \times 216 \times 14 \times 14\right] \times 11=$  & $931{,}392$ \\
    \midrule
    res$_5$         & residual$_1$          & $(3-1-1) \times 96 \times 14 \times 14 =$                       & $18{,}816$ \\
                    & residual$_{2-3}$      & $\left[(3-1-1) \times 192 \times 7 \times 7\right] \times 6 =$  & $56{,}448$ \\
                    & conv$_{1-3}$          & $\left[(3-1) \times 432 \times 7 \times 7\right] \times 7 =$  & $296{,}352$ \\
    \midrule
    pool$_5$        & -                     & $(16-1) \times 432 =$                                         & $6{,}480$ \\
    \bottomrule
    \textbf{Total}  &                       &                                                               & \textbf{4,771,632} \\ 
    \bottomrule
\end{tabular}
\end{center}
\caption{\textbf{\textit{Co}X3D-M} state memory consumption by layer. }
\label{tab:cox3dm-mem-calc}
\end{table}

\begin{table}[!htbp]
\begin{center}
\begin{tabular}{lcc}
    \toprule
    \multirow{2}{*}{\textbf{Stage}} & \multirow{2}{*}{\textbf{Filters}}   & \textbf{Output size} \\ 
                                    &                                     & ($T \times H \times W$) \\
    \midrule
    input           & -                         & $16 \times 224 \times 224$ \\
    \midrule
    conv$_1$        & $\begin{matrix} 
                      1 \times 3^2, 24 \\
                      \mathcolorbox{greycol}{5^{*} \times 1^2, 24}
                      \end{matrix}$             & $16 \times 112 \times 112 $ \\
    \midrule
    res$_2$         & \highlight{res}
                      $\begin{bmatrix} 
                      1 \times 1^2, 54 \\
                      \mathcolorbox{greycol}{\ 3 \times 3^2, 54 \ } \\
                      \text{SE} \\
                      1 \times 1^2, 24 \\
                      \end{bmatrix} \times 3$   &$16 \times 56 \times 56 $ \\
    \midrule
    res$_3$         & \highlight{res}
                      $\begin{bmatrix} 
                      1 \times 1^2, 108 \\
                      \mathcolorbox{greycol}{3 \times 3^2, 108} \\
                      \text{SE} \\
                      1 \times 1^2, 48 \\
                      \end{bmatrix} \times 5$   &$16 \times 28 \times 28 $ \\
    \midrule
    res$_4$         & \highlight{res}
                      $\begin{bmatrix} 
                      1 \times 1^2, 216 \\
                      \mathcolorbox{greycol}{3 \times 3^2, 216} \\
                      \text{SE} \\
                      1 \times 1^2, 96 \\
                      \end{bmatrix} \times 11$   &$16 \times 14 \times 14 $ \\
    \midrule
    res$_5$         & \highlight{res}
                      $\begin{bmatrix} 
                      1 \times 1^2, 432 \\
                      \mathcolorbox{greycol}{3 \times 3^2, 432} \\
                      \text{SE} \\
                      1 \times 1^2, 192 \\
                      \end{bmatrix} \times 7$   &$16 \times 7 \times 7 $ \\
    \midrule
    conv$_5$        & $1 \times 1^2, 432 $      & $16 \times 7 \times 7 $ \\
    pool$_5$        & $\mathcolorbox{greycol}{16 \times 7^2}$  & $1 \times 1 \times 1$ \\
    fc$_1$          & $1 \times 1^2, 2048 $     & $1 \times 1 \times 1 $ \\
    fc$_2$          & $1 \times 1^2, \text{\#classes} $ & $1 \times 1 \times 1 $ \\
    \bottomrule
\end{tabular}
\end{center}
\caption{\textbf{X3D-M model architecture}. When converted to a continual CNN, the \highlight{highlighted} components carry an internal state which results in a memory overhead. *Temporal kernel size in conv$_1$ is set to 5 as found in the official X3D source code~\cite{feichtenhofer2020x3d}.}
\label{tab:x3d-mem}
\end{table}

\clearpage
\section{Benchmarking details} \label{apx:benchmark-details}
This section should be read in conjunction with \cref{sec:inference-benchmarks} of the main paper.
To gauge the achievable on-hardware speeds of clip and frame predictions, a benchmark was performed on the following four system:
A CPU core of a MacBook Pro (16-inch 2019 2.6 GHz Intel Core i7); 
Nvidia Jetson TX2;
Nvidia Jetson Xavier;
and a Nvidia RTX 2080 Ti GPU (on server with Intel XEON Gold processors).
A batch size of 1 was used for testing on CPU, while the largest fitting multiple of $2^\mathbb{N}$ up to 64 was used for the other hardware platforms which have GPUs and lend themselves better to parallelisation. 
Thus, the speeds noted for GPU platforms in \cref{tab:benchmark-kinetics400} of the main paper should not be interpreted as the number of processed clips/frames from a single (high-speed) video stream, but rather as the aggregated number of clips/frames from multiple streams using the available hardware.
The exact batch size and input resolutions can be found in \cref{tab:benchmark-config}. 
In conducting the measurements, we assume the input data is readily available on the CPU and measure the time it takes for it to transfer from the CPU to GPU (if applicable), process, and transfer back to the CPU.
A precision of 16 bits was used for the embedded platforms TX2 and Xavier, while a 32 bit precision was employed for CPU and RTX 2080 Ti. 
All networks were implemented and tested using PyTorch, and neither Nvidia TensorRT nor ONNX Runtime were used to speed up inference.


\begin{table}[!htbp]
\begin{center}
\begin{tabular}{lrrrrr}
    \toprule
    \textbf{Model} 
    &\textbf{Input shape} 
    &\multicolumn{4}{c}{\textbf{Batch size}}
        \\ \cline{3-6}
        &($T \times S^2$) &\textbf{CPU} & \textbf{TX2}  &\textbf{Xavier} &\textbf{RTX}
    \\
    \midrule
    
    I3D-R50                       & $8  \times 224^2 $      & 1       & 16      & 16       & 32     \\
    R(2+1)D-18$_8$                & $8  \times 112^2 $      & 1       & 16      & 16       & 32     \\
    R(2+1)D-18$_{16}$             & $16 \times 112^2 $      & 1       & 8       & 16       & 32     \\
    Slow-8×8-R50                  & $8  \times 256^2 $      & 1       & 8       & 8        & 8     \\ 
    SlowFast-8×8-R50              & $8  \times 256^2 $      & 1       & 8       & 32       & 32     \\ 
    SlowFast-4×16-R50             & $16 \times 256^2 $      & 1       & 16      & 32       & 32     \\ 
    X3D-L                         & $16 \times 312^2 $      & 1       & 16       & 32       & 32     \\
    X3D-M                         & $16 \times 224^2 $      & 1       & 32      & 64       & 64     \\
    X3D-S                         & $13 \times 160^2 $      & 1       & 64      & 64       & 64     \\
    X3D-XS                        & $4  \times 160^2 $      & 1       & 64      & 64       & 64     \\
    \textit{Co}I3D                & $1  \times 224^2 $      & 1       & 8       & 8        & 8     \\
    \textit{Co}Slow                & $1  \times 224^2 $      & 1       & 8       & 8        & 8     \\
    \textit{Co}X3D-L              & $1  \times 312^2 $      & 1       & 8       & 16       & 32     \\
    \textit{Co}X3D-M              & $1  \times 224^2 $      & 1       & 32      & 64       & 64     \\
    \textit{Co}X3D-S              & $1  \times 160^2 $      & 1       & 32      & 64       & 64     \\
    \bottomrule
\end{tabular}
\end{center}
\caption{
    \textbf{Benchmark model configurations}. For each model, the input shape is noted as $T \times S^2$, where $T$ and $S$ are the temporal and spatial input shape. 
}
\label{tab:benchmark-config}
\end{table}

\begin{table}
\begin{center}
\begin{tabular}{llccccccc}
    \toprule
    \multirow{2}{*}{\textbf{Model}} 
    &\multirow{2}{*}{\textbf{FLOPs}}  
    &\multicolumn{4}{c}{\textbf{Throughput (evaluations/s)}}
        \\ \cline{3-6}
        &&CPU & TX2  &Xavier &RTX
    \\
    \midrule             
    (\textit{Co})I3D     &5.04$\times$   & 3.39$\times$  & 0.95$\times$  & 1.62$\times$  & 1.64$\times$ \\
    (\textit{Co})Slow    &7.95$\times$   & 7.68$\times$  & 1.19$\times$  & 1.44$\times$  & 1.65$\times$ \\
    (\textit{Co})X3D-L   &15.34$\times$  & 9.20$\times$  & 5.21$\times$  & 5.77$\times$  & 5.98$\times$ \\
    (\textit{Co})X3D-M   &15.06$\times$  & 9.05$\times$  & 4.79$\times$  & 4.95$\times$  & 6.86$\times$ \\
    (\textit{Co})X3D-S   &12.11$\times$  & 5.91$\times$  & 4.15$\times$  & 4.98$\times$  & 3.41$\times$ \\
    \bottomrule
\end{tabular}
\end{center}
\caption{\textbf{Relative improvements} in frame-by-frame inference in Continual 3D CNN relative to regular 3D CNN counterparts. The improvements ($\times$ lower for FLOPs and $\times$ higher for throughput) correspond to the results in \cref{tab:benchmark-kinetics400} of the main paper.}
\label{tab:relative-results}
\end{table}


\section{A note on RCU FLOPs} \label{apx:rcu}
In \cref{tab:benchmark-kinetics400} of the main paper, we have approximated the FLOPs for RCU~\cite{singh2019recurrent} as follows:
We use a different measure of FLOPs (the one from the \texttt{ptflops}~\cite{sovrasov2020ptflops}) than the RCU authors and therefore employ a translation factor of $28.6 / 41.0$, which is our measured FLOPs for I3D ($28.6$) divided by theirs ($41.0$), multiplied with their reported $54.0$ for RCU. Considering that their method used 8 frames and can be applied per frame, we also divide by 8. 
Note that the this approximation lacks the repeat classification layer and may thus be considered on the low side. The resulting computation becomes 
$28.6 / 41.0 \cdot 54.0 / 8 = 4.71$.


\section{Supplemental visualisations of benchmark}
\begin{figure}[b]
    \centering
    \includegraphics[width=0.8\linewidth]{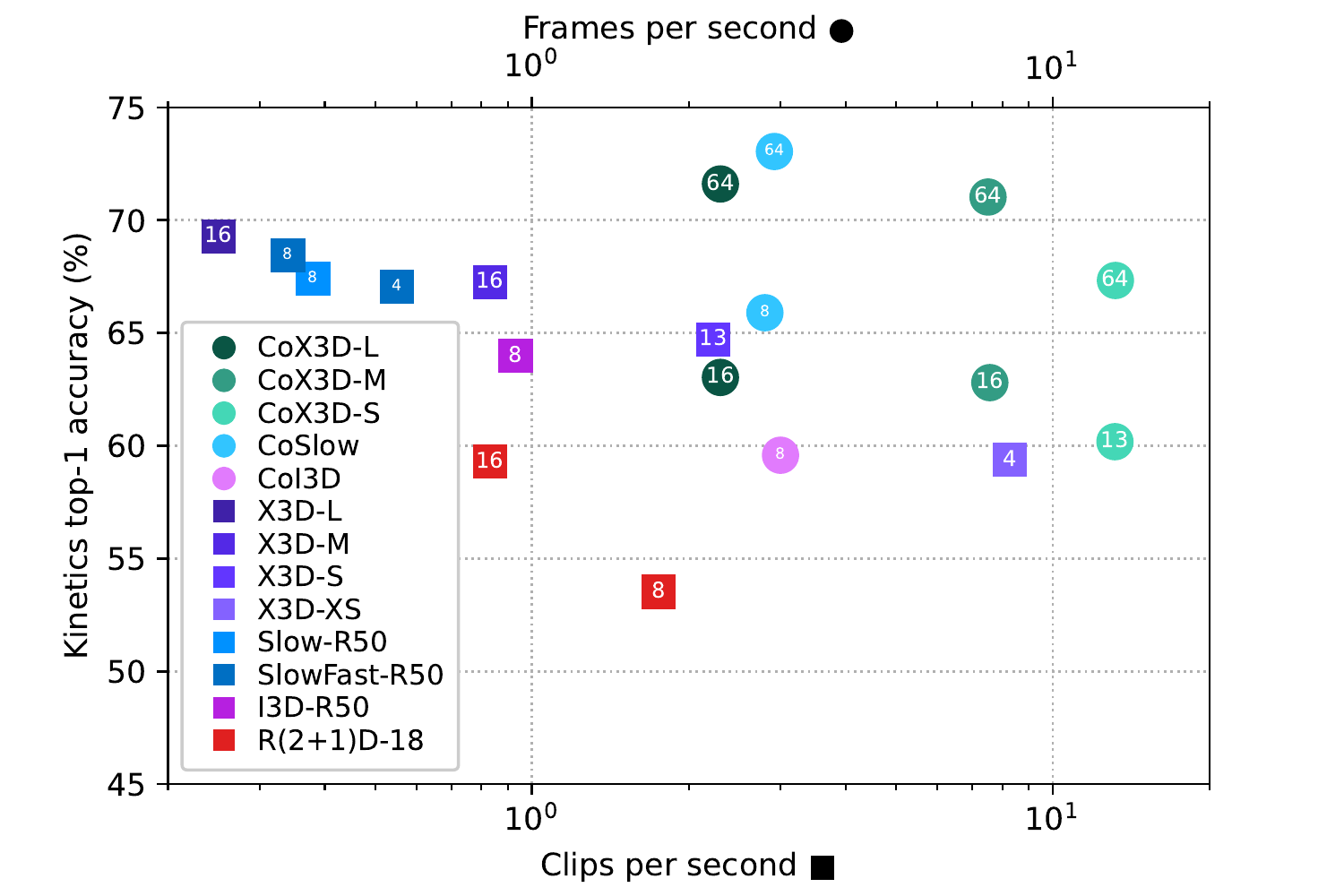}
    \caption{\textbf{CPU} inference throughput versus top-1 accuracy on Kinetics-400. } 
    \label{fig:acc-vs-speed-cpu}
\end{figure}

\begin{figure}[b]
    \centering
    \includegraphics[width=0.8\linewidth]{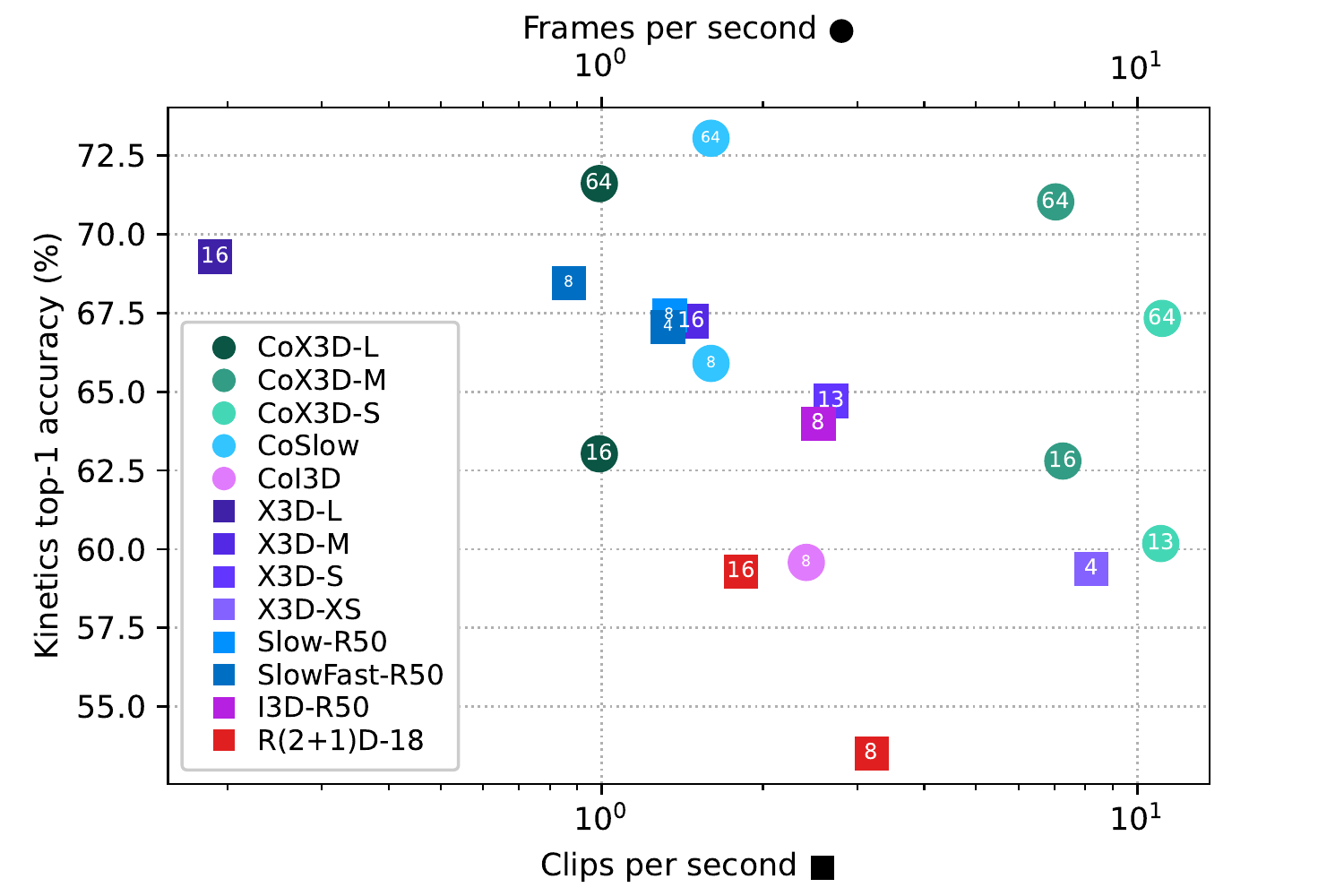}
    \caption{\textbf{TX2} inference throughput versus top-1 accuracy on Kinetics-400. }
    \label{fig:acc-vs-speed-tx2}
\end{figure}

\begin{figure}[b]
    \centering
    \includegraphics[width=0.8\linewidth]{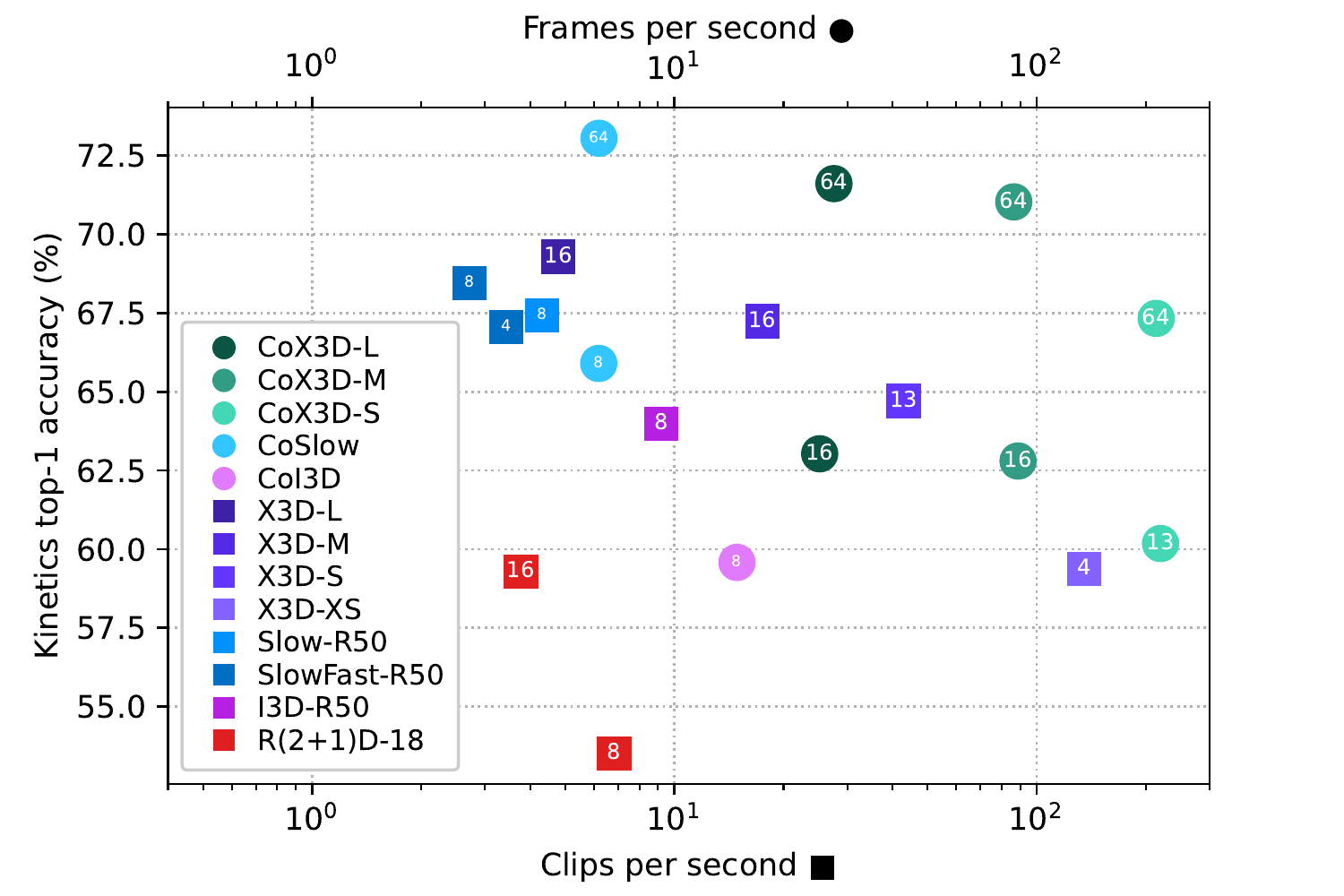}
    \caption{\textbf{Xavier} inference throughput versus top-1 accuracy on Kinetics-400.}
    \label{fig:acc-vs-speed-xavier}
\end{figure}

\begin{figure}[!b]
    \centering
    \includegraphics[width=0.8\linewidth]{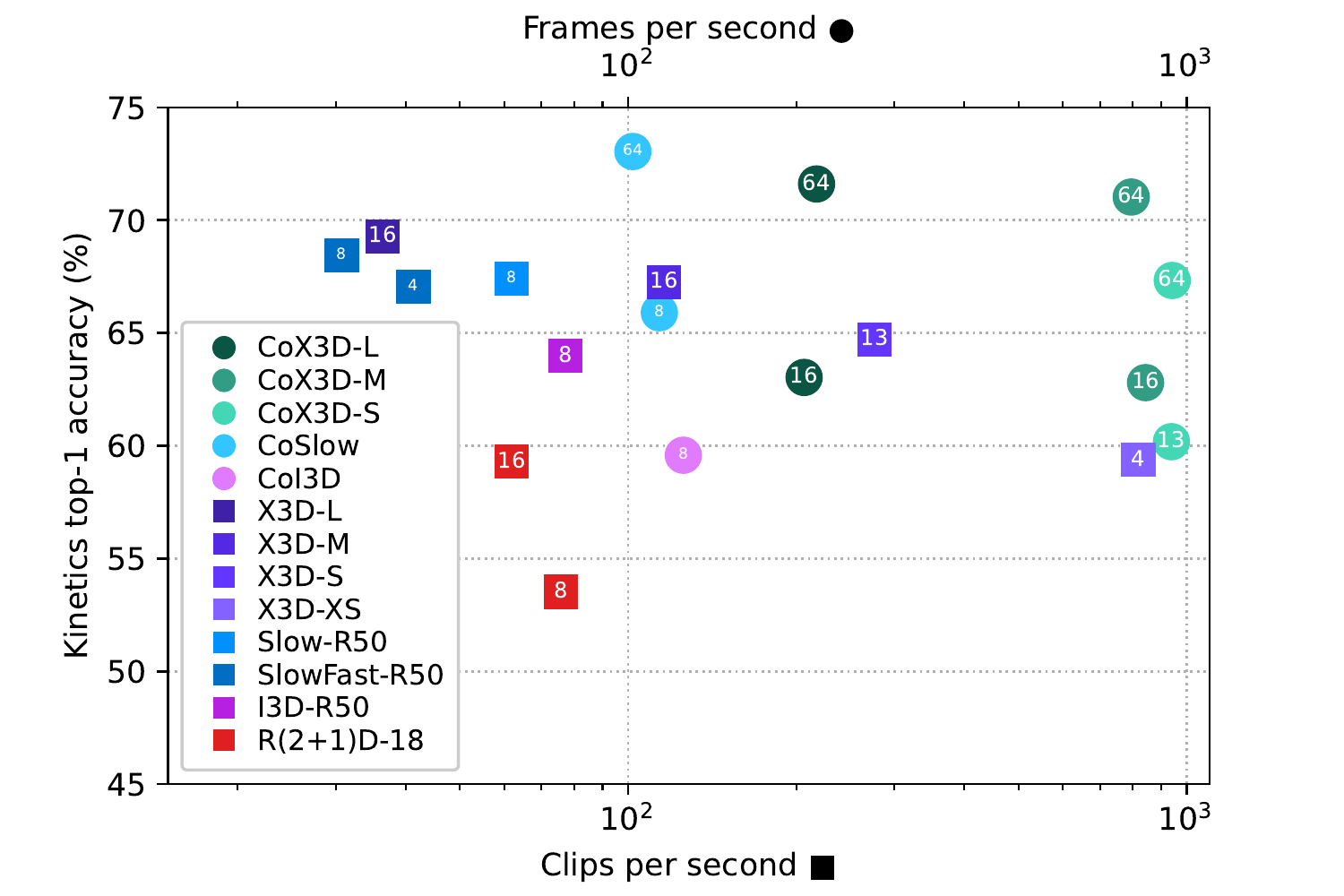}
    \caption{\textbf{RTX2080Ti} inference throughput versus top-1 acc. on Kinetics-400.}
    \label{fig:acc-vs-speed-rtx}
\end{figure}

As a supplement to the results presented in the main paper, this appendix supplies additional views of the benchmarking results in \cref{tab:benchmark-kinetics400}.
Accordingly, graphical representations of the accuracy versus speed trade-offs from \cref{tab:benchmark-kinetics400} are shown in Figures \ref{fig:acc-vs-speed-cpu}-\ref{fig:acc-vs-speed-rtx}.
As in \cref{fig:test-acc-vs-flops} of the main paper, the noted accuracies on Kinetics-400 were achieved using 1-clip/frame testing on publicly available pretrained models, the \textit{Co}X3D models utilised X3D weights without further fine-tuning, and the numbers noted in each point represent the size of the global average pooling layer.
Likewise, \cref{tab:relative-results} shows the improvements in continual inference relative to the regular models.
In general, the FLOPs improvements are higher than on-hardware speed evaluations, with relatively lower improvements on hardware platforms with GPUs. 
We attribute these differences to a memory operations overhead, which does not enjoy the same computational improvement as multiply-accumulate operations do on massively parallel hardware.

From Figures \ref{fig:acc-vs-speed-cpu}-\ref{fig:acc-vs-speed-rtx} we likewise observe, that the I3D, R(2+1)D and SlowFast models perform relatively better on hardware compared to the X3D and \textit{Co}X3D models, which utilise computation-saving approaches such as 1D-convolutions and grouped 3D-convolutions at the price of increasing memory access cost.

\end{document}